# Thinking Less to Simulate Better: Intuitive Prompting Improves LLM Agents Simulating Individual Social Media Reactions, Including Unfamiliar Content

Ljubisa Bojic[1,2,3,*], Ph. D., Senior Research Associate
(Corresponding author; Email address: ljubisa.bojic@ivi.ac.rs; ORCID: 0000-0002-5371-7975)

Tijana Stanic[4], Master Student
(Email address: tijana.s01@gmail.com; ORCID: https://orcid.org/0009-0003-6000-1857

Jörg Matthes[5], Ph. D., Full Professor,
(Email address: joerg.matthes@univie.ac.at; ORCID: 0000-0001-9408-955X)

Agariadne Dwinggo Samala[6], Ph. D., Associate Professor
(Email address: agariadne@ft.unp.ac.id; ORCID: 0000-0002-4425-0605)

Bojana Dinic[7], Ph. D., Full Professor
(Email address: bojana.dinic@ff.uns.ac.rs; ORCID: 0000-0002-5492-2188)

Jue Wang[8], Ph. D., Associate Professor
(Email address: wangjue@ntu.edu.sg; ORCID: 0000-0002-3401-713X)

[1]Institute for Artificial Intelligence Research and Development of Serbia; Address of correspondence: 1 Frukogorska 1, 21000 Novi Sad, Serbia;
[2]University of Belgrade, Institute for Philosophy and Social Theory, Digital Society Lab; Address of correspondence: Kraljice Natalije 45, 11000 Belgrade, Serbia;
[3]Complexity Science Hub, Vienna, Austria; Address of correspondence: Metternichgasse 8, 1030 Vienna, Austria;
[4] University of Belgrade, Faculty of Political Sciences, Kragujevac, Serbia; Address of correspondence: Jove Ilica 165, 11000 Belgrade, Serbia;
[5] University of Vienna, Faculty of Social Sciences, Department of Communication; Address of correspondence: Währinger Straße 29 (R. 7.47), 1090 Vienna, Austria;
[6] Universitas Negeri Padang, Faculty of Engineering, Padang, Indonesia; Address of correspondence: 482X+WQ West Air Tawar, Padang City, West Sumatra, Indonesia;
[7]University of Novi Sad, Faculty of Philosophy, Novi Sad, Serbia; Address of correspondence: Dr Zorana Đinđića 2, 21102 Novi Sad, Serbia;
[8] Nanyang Technological University, School of Social Sciences, Singapore; Address of correspondence: 50 Nanyang Ave, 639798 Singapore;

## Abstract

Platform policies are increasingly tested on artificial users, making agent fidelity important. Yet convincing fake profiles could also manipulate perceived public opinion before elections. Validation has concentrated on agreement with human behaviour and has paid little attention to whether an agent behaves in line with the profile it was given. The present study profiled eight Serbian participants through a questionnaire, a deep interview, and a written self-presentation, recorded their reactions to sixty-eight social media posts, and asked four language models to predict those reactions under five prompt conditions varying profile content and instruction style. Attitudinal content improved prediction over demographic backstories by a wide margin. Agents matched their stated profiles more closely than participants matched their own survey answers, and consistency proved unrelated to fidelity once profile information was present. Instructing models to respond intuitively and immediately rather than analytically gave the highest fidelity of any condition and cut the compression of individual differences from seven times the human level to three. The advantage held on posts about topics the questionnaire never raised, where that condition reached the highest fidelity of any setup and beat a crowd baseline by a wide margin, which suggests that agents prompted this way could serve as general-purpose simulated users rather than specialists on the topics they were profiled for. Results may bear implications for the development of language models, because intuition-based setups appear better suited to some tasks than reasoning-based ones.

## 1. Introduction

Social media platforms shape what hundreds of millions of people read, believe, and argue about every day, and the systems that govern them are increasingly tested on simulated audiences before they reach real ones. Large language models make it possible to create artificial users that carry the attributes of specific people and react to content on their behalf. The same capability carries an obvious danger. Fake profiles built this way cost almost nothing to produce and can be deployed in large numbers, so a swarm of accounts that behave like plausible local citizens could enter the information ecosystem of a country and remain there for months without drawing attention. Coordinated networks of that kind would shift the apparent balance of opinion before an election while appearing to consist of ordinary users, and experimental work already shows that conversations with language models move voter preferences by margins larger than those achieved by conventional political advertising (Hackenburg et al., 2025). The technical problem behind the threat and the technical problem behind the research tool are the same one, since both depend on how faithfully a model can act as a particular person. Agents that reproduce individual reactions well would let researchers pilot platform policies, moderation rules, and public communication campaigns without exposing anyone to the risk of the experiment. Failure would be harder to detect than success, because a simulation can produce populations that look plausible while quietly erasing the differences between the people in them. Fidelity at the level of the individual is therefore a question of governance rather than a technical curiosity, and it becomes more pressing as recommender systems and generative models take on a larger role in shaping what people encounter online (Bojic, 2024).

Research on artificial respondents began with the observation that language models conditioned on demographic backstories can approximate the response distributions of the groups those backstories describe. Argyle et al. (2023) introduced the idea of algorithmic fidelity and showed that a model prompted with American sociodemographic profiles reproduced several patterns found in national election studies. Aher et al. (2023) extended the approach by replicating classic experiments in economics and psychology with simulated participants, and reported that well known treatment effects appeared in the artificial data. Grossmann et al. (2023) and Dillion et al. (2023) discussed what this would mean for the social sciences, with the second paper cautioning that agreement at the level of averages says little about whether a model captures the reasoning of any particular person. Park et al. (2023) moved from survey responses to behaviour by building generative agents that planned, remembered, and interacted inside a small simulated town, and a later study extended the method to interview-based replicas of more than a thousand people (Park et al., 2024). Larger platforms followed, including OASIS with a million agents (Yang et al., 2024) and AgentSociety (Piao et al., 2025), alongside work that used simulated feeds to compare alternative ranking algorithms (Törnberg et al., 2023) and to trace opinion dynamics in networks of agents (Chuang et al., 2023). Agent-based designs have also been applied to politically charged content, where personalisation and polarisation can be observed under controlled conditions (Bojic et al., 2025a).

Evidence that models reflect some populations better than others accumulated quickly. Santurkar et al. (2023) compared model outputs against sixty American demographic groups and found misalignment on a scale comparable to the gap between the two main political parties. Durmus et al. (2024) reported similar imbalances across countries, with responses tilted toward the views of wealthy Western populations. Cheng et al. (2023) showed that persona prompts push models toward caricature, producing exaggerated versions of the groups they are meant to represent, and Zewail et al. (2026) documented the same tendency in moral judgment. Guilbeault et al. (2025) traced distortions of age and gender that appear in both online media and the models trained on it. These results share a structure worth noting. Models perform respectably when asked about aggregates and much less well when asked about the particular case, which is why several authors now argue that simulation is safest as a tool for discovering collective patterns rather than individual trajectories (Wu et al., 2025). The imbalance carries a further implication for research conducted outside the English-speaking world, since a model that reflects wealthy Western publics will misjudge a Serbian sample in ways that are systematic rather than random.

A parallel line of work asked how much a persona actually changes model behaviour. Hu and Collier (2024) quantified the persona effect in subjective annotation tasks and found that persona variables explained only a small share of variance, even though persona prompts did shift predictions in measurable ways. Bodroza et al. (2024) administered personality inventories to language models and reported limited temporal stability together with a consistent pull toward prosocial answers. Serapio-García et al. (2025) built a psychometric framework for shaping and measuring traits in models, and showed that traits can be induced reliably while remaining sensitive to how the prompt is written. Work on inference in the opposite direction has been more encouraging, with models able to recover psychological dispositions from social media text (Peters & Matz, 2024) and from written material more generally (Derner et al., 2024), which follows a longer tradition of predicting private attributes from digital traces (Kosinski et al., 2013). The asymmetry is striking. Reading a person from their words appears easier than writing that person forward into new situations.

Capability assessments that avoid personas altogether suggest the underlying language competence is considerable. Comparisons between models and human annotators on sentiment, political leaning, emotional intensity, and sarcasm show that current systems approach human levels on several of these tasks while diverging on others (Bojic et al., 2025c), and tests of pragmatic understanding point in a similar direction (Bojic et al., 2025b). Mahowald et al. (2024) offered a framework for reading such results by separating formal linguistic competence from the functional thinking that many tasks require. The distinction matters for simulation research, because a model that understands a post perfectly well may still misjudge how a specific reader would respond to it. Competence with language and fidelity to a person are separate achievements, and progress on the first does little to guarantee the second.

The most consistent negative finding across this literature concerns variance. Synthetic respondents cluster more tightly than real ones, a pattern described as under-dispersion or compression of individual differences. Xie et al. (2026) evaluated the statistical realism of model-generated social science data and found that compressed variance can flip regression signs and manufacture significant effects where human data show none. Dominguez-Olmedo et al. (2024) raised related concerns about the survey responses of language models, including sensitivity to question format and answer ordering. Compression matters for reasons that reach past statistics, since a simulated public whose members resemble one another will understate disagreement and make minority positions disappear from any policy test run on it. The concern connects to broader work on value alignment between machines and the societies they serve, where the risk lies in optimising against a picture of human preferences that has already been smoothed (Bojic et al., 2026).

Validation practice has struggled to keep pace with the volume of simulation studies. Münker et al. (2025) argued that generative agents should be benchmarked against empirical realism before being trusted to mimic communication on social networks, and Schwager et al. (2026) developed operational validity tests for conditioned comment prediction. Loru et al. (2025) examined the simulation of judgment and found systematic differences between model and human evaluation. Most of these efforts measure agreement between agent output and human behaviour, which is the natural first question. Recent methodological guidance for psychology argues that the validity requirements should scale with the ambition of the claim, so that using a model to code text demands less than treating it as a stand-in for a person (Lin, 2026).A second question has received far less attention. An agent given a profile can be assessed against that profile, by asking whether its reactions follow from the attitudes it was told the person holds. Social psychology has long known that people themselves score poorly on this test, with a gap between stated attitudes and observed behaviour reported since LaPiere (1934) and formalised in later reviews (Wicker, 1969; Ajzen & Fishbein, 1977). An agent that matches its profile perfectly would therefore be behaving in a way no real person does, which turns internal consistency into a diagnostic rather than a goal.

How a model is asked to think about a profile may matter as much as what the profile contains. Instructions that invite deliberate reasoning improve performance on many benchmarks, and the competence distinction drawn by Mahowald et al. (2024) helps explain why such gains appear on some tasks and vanish on others. Human reactions to social media posts belong to a family of judgments made in a second or two, driven by immediate response rather than by weighing evidence, in the sense described by dual process accounts of cognition (Kahneman, 2011). A model instructed to reason its way from an attribute table to an answer may therefore be solving the wrong problem, and may drift toward the stereotype of a person who holds those attributes rather than toward the person. The design used here tests that possibility directly by varying both the content of the profile and the style of the instruction across five conditions.

Sample size deserves a word, since the design departs from the large panels common in this field. Survey satisficing, where respondents give acceptable rather than accurate answers to reduce effort, has been documented for decades (Krosnick, 1991), and its effects are strongest in long online instruments of the kind used to build synthetic populations. A small number of participants profiled through structured questionnaires, deep interviews, and free-text self-presentation yields ground truth of a quality that large low-engagement panels rarely reach. The trade is deliberate, favouring depth per person over breadth across people, which suits a study whose central questions concern individuals rather than population estimates. Choices of this kind also carry a methodological cost that should be stated plainly, since findings from eight people describe what happens within a rich-profile paradigm rather than what would hold across a national population.

The present study profiles eight Serbian participants in depth, records their reactions to sixty-eight social media posts, and asks four language models to predict those reactions under five prompt conditions that vary profile content and instruction style. Fidelity is reported as the Matthews correlation coefficient, which uses all four cells of the confusion matrix and resists the inflation that accuracy suffers under class imbalance (Chicco & Jurman, 2020). Ten hypotheses follow from the literature reviewed above. H1 holds that profiled agents predict individual reactions above chance. H2 holds that attitudinal survey content improves prediction substantially over demographic information alone. H3 holds that language model agents outperform conventional supervised models trained on the same survey variables. H4 holds that returns to qualitative profile content are small once a structured questionnaire is available. H5 holds that agents are more consistent with a stated profile than the people described by that profile. H6 holds that consistency and fidelity are distinct properties whose relationship is non-monotonic, rising together when profile information is absent and diverging once it is supplied. H7 holds that instructions which suppress deliberation improve fidelity and reduce the compression of individual differences. H8 holds that instructions which invite explicit inference from combinations of attributes fail to improve fidelity. H9 holds that agents generalise to topics the profile never covered, with fidelity above chance yet well below the level reached on profiled topics. H10 holds that narrative profile content supports this generalisation in a way that tabulated attributes do not. H11 holds that intuitive framing improves generalisation to unprofiled topics, which would indicate that agents built this way can act as general-purpose simulated users.

## 2. Methodology

The study combines a small qualitative panel with a factorial computational experiment. Participants were profiled in depth and recorded their own reactions to a fixed set of social media posts, after which four language models predicted those same reactions under five prompt conditions. The design allows three properties of each agent to be measured against the same human data, namely fidelity to what people actually did, consistency with what people said about themselves, and the degree to which agents differentiate between individuals. Figure 1 shows the sequence of stages. All data, code and reproducibility instructions can be accessed publicly (OSF, 2026).

### 2.1 Participants

Eight adults living in Serbia took part. Ages ranged from 22 to 53 years with a mean of 31.0, and the panel included five women and three men. Educational attainment spanned current undergraduate enrolment through completed master studies, and employment status covered students, employed professionals, and freelance workers. Recruitment aimed for variation in political self-placement and institutional trust rather than demographic representativeness, since the analysis concerns prediction of specific individuals rather than estimation of population parameters. Panels of this size fall within the range where thematic saturation is commonly reached in interview research (Guest et al., 2006), and the depth of profiling per person is the resource the design depends on.

Two participants from an initial group of ten were withdrawn before the computational stage because their profiles were incomplete. Participant identifiers run from P01 to P10 with P02 and P04 absent, and the identifiers were kept unchanged so that the data files remain traceable to the original collection.

### 2.2 Profile construction

Each participant completed a questionnaire administered in Serbian, covering demographics, social media use, news sources, frequency of AI chatbot use, political interest, left-right self-placement, two attitude statements on leadership and sovereignty, trust in eleven Serbian institutions on a five-point scale, eleven personality and values statements on a five-point scale, belief in astrology, religious affiliation, and a multiple-choice item on preferred leisure activities. An attention check was embedded in the instrument. The questionnaire was written in the participants' own language to avoid the translation effects that complicate cross-national instruments (Smith, 2003).

A semi-structured interview followed the questionnaire and explored daily routines, media habits, opinions on current events, and personal history in the participant's own words. Each participant also wrote a short free-text self-presentation in response to an open prompt asking how they would introduce themselves at a party. The interview transcripts and the self-presentations supply the narrative material that distinguishes the richer prompt conditions from the tabulated ones. Analysis of the interview content followed standard practice for organising qualitative material into themes (Braun & Clarke, 2006), and the transcripts entered the prompts in full rather than in summarised form.

## 2.3 Stimulus set

Sixty-eight posts written in Serbian formed the stimulus set. Thirty-four posts address topics covered by the questionnaire and thirty-four address topics the questionnaire never raised, which allows in-domain and out-of-domain prediction to be separated. Thirty-eight posts concern news and politics and thirty concern entertainment and lifestyle. Related posts come in matched pairs, with one post expressing a positive stance on a topic and the other expressing a negative stance, producing seventeen topic pairs. Nine pairs address trust in institutions, seven address leisure activities, and one addresses artificial intelligence. Six additional posts state political positions on contested issues such as the student protests, Israel and Palestine, and Russia and Ukraine, and these have no matched counterpart.

A post key records for every post its identifier, its valence, whether it relates to the survey, and the specific survey item it corresponds to. The key was prepared during stimulus construction rather than after data collection, which matters because the consistency measure depends on it. Posts were presented to participants in Serbian and to the models in English translation, with the translation verified against the original text for every item.

Participants reacted to each post by selecting like or dislike, with the two options mutually exclusive. The procedure yielded 544 human reactions across the eight participants. Overall the panel liked 51.1 per cent of posts, which leaves the two response categories close to balanced.

## 2.4 Prompt conditions

Four language models were tested, namely Claude Sonnet 5, DeepSeek Instant, GPT 5.6 Terra, and Grok Fast. Every model received every participant profile under five prompt conditions, and each prompt asked for a like or dislike judgment on all sixty-eight posts returned in a fixed comma-separated format. The five conditions form a ladder that varies profile content across the first three levels and instruction style across the last two.

The demographics condition, labelled V4, supplied age, gender, education, employment status, and city of residence. The attributes condition, labelled V3, added the full questionnaire including the trust battery, the personality items, the activity list, and political self-placement. The interview condition, labelled V1, added the interview transcript and the free-text self-presentation, which makes it the richest profile in the study. The inference condition, labelled V2, used the same content as V1 together with an instruction telling the model to infer reactions from combinations of characteristics when a directly relevant attribute was absent. The intuitive condition, labelled V5, also used the V1 content but instructed the model to respond immediately rather than analytically, to avoid weighing characteristics against one another, and to accept that some reactions would sit at odds with the stated profile.

The design produced twenty configurations and 10,880 predicted reactions. Two further models were tested and excluded because their output files were incomplete, and the exclusion was made before any analysis of their content. All model outputs were converted to a single schema of post identifier, agent identifier, like, and dislike, then checked for row count, participant coverage, duplicate entries, and contradictory values. Every retained file contained 544 valid rows.

## 2.5 Outcome measures

Fidelity is agreement between an agent’s prediction and the participant’s actual reaction. The Matthews correlation coefficient serves as the primary measure because it draws on all four cells of the confusion matrix and stays informative when one response category dominates, a property that accuracy and several related indices lack (Chicco et al., 2021). Accuracy, Cohen’s kappa (Cohen, 1960), and the F1 score are reported alongside it for readers who prefer familiar scales, with kappa values interpreted against conventional benchmarks (Landis & Koch, 1977).

Profile-consistency asks a different question. For each topic pair, the participant's own survey answer implies an expected pattern of reactions, and the measure records whether an actor produced that pattern. Trust scores of four or five imply agreement with the positive post and disagreement with the negative one, trust scores of one or two imply the reverse, and a neutral score of three is treated as non-diagnostic and excluded. Activity items are handled symmetrically according to whether the participant listed the activity. Frequency of AI use is split at the same point on its six-point scale. The rule set is implemented in code with all thresholds exposed as parameters, which allows the sensitivity of the results to these choices to be examined directly.

Scoring proceeds pair by pair. A full match requires both posts in a pair to align with the expected pattern, and a partial score gives half credit when one post aligns. The measure is defined identically for the human participant and for every agent, which makes the two directly comparable on the same scale. Application of the rules to eight participants and seventeen topics gave 115 evaluable cells per actor, with 21 cells excluded because the underlying survey answer was neutral.

Individuation is the standard deviation across participants of an actor's overall like-rate. Comparison of this quantity for agents against the same quantity for the human panel gives a compression ratio, where values above one indicate that an agent treats different people more alike than they actually are.

### 2.6 Baselines and statistical procedures

Agent performance was compared against five baselines of increasing strength. The majority-class baseline predicts the more common response throughout. The participant-prior baseline predicts each person's own modal reaction. The crowd baseline predicts each post's modal reaction among the other seven participants, which represents what can be achieved with no knowledge of the individual. Two supervised models complete the set, namely logistic regression and a random forest (Breiman, 2001), each trained on the full set of survey variables together with post attributes and fitted with scikit-learn (Pedregosa et al., 2011).

All baselines were evaluated under leave-one-participant-out cross-validation, so the held-out person never appears in training. The procedure matches the position of an agent handed a profile it has never seen, and it avoids the optimistic estimates that in-sample evaluation would produce at this sample size (Varoquaux, 2018).

Comparisons between conditions used McNemar's test on paired agent-post observations (McNemar, 1947), which treats each of the 544 cells as a unit and is well powered for detecting differences between two prediction sets on the same data. Comparisons treating the model as the unit of analysis used the Wilcoxon signed-rank test across the four models, and results from both are reported because they answer different questions and carry very different power. Associations between consistency and fidelity across configurations were assessed with Pearson correlation, calculated across all twenty configurations and again across the sixteen configurations that received attitudinal profile information.

### 2.7 Ethics and reproducibility

Participants gave informed consent covering the questionnaire, the interview, and the use of their profiles to condition language models. Identifiers replace names throughout, and the interview transcripts were checked for details that might identify a person before entering any prompt. All analysis code, the prompt templates, the post key, and the converted model outputs are available so that the rule thresholds can be varied and the analysis repeated.

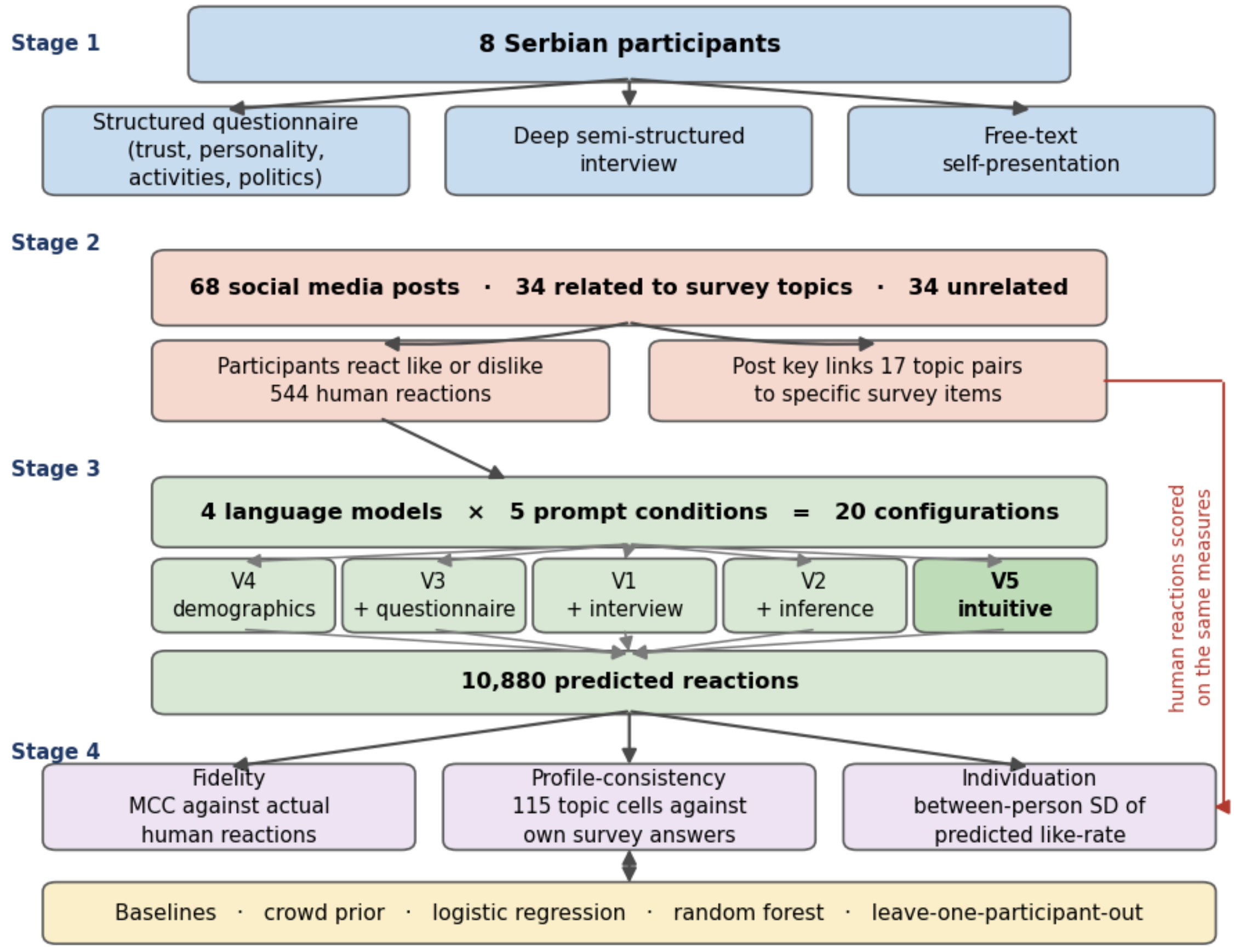

*Figure 1. Stages of the research design.*

## 3. Results

The eight participants liked 51.1 per cent of the sixty-eight posts, which leaves the two response categories close to balanced and keeps chance performance near zero on the Matthews correlation coefficient. Agreement among participants varied widely across the stimulus set, with 33 of the 68 posts drawing agreement from six or fewer of the eight people. Posts of that kind carry the individual-level signal the study aims to capture, since a simulation that treats everyone alike can succeed on the remaining items without representing anyone.

### 3.1 Fidelity across prompt conditions

Every one of the twenty configurations predicted individual reactions above chance, with the Matthews correlation coefficient ranging from 0.275 for Claude Sonnet 5 given demographics alone to 0.511 for Grok Fast under the interview and intuitive conditions. The result supports H1 and places all configurations in territory where the predictions carry real information about the person, since values in this range indicate moderate association on a measure that stays honest under class imbalance (Boughorbel et al., 2017).

Table 1 reports each outcome averaged across the four models. Fidelity climbs from the demographics condition to the questionnaire condition, gains a little more from the interview material, stays flat when the inference instruction is added, and reaches its highest value under intuitive framing.

*Table 1. Outcome measures by prompt condition, averaged across the four models.*

| Condition | MCC | Accuracy | Consistency | MCC contested | Between-person SD | Compression |
|---|---|---|---|---|---|---|
| V4 demographics | 0.300 | 0.650 | 0.541 | 0.291 | 0.0053 | 10.24× |
| V3 questionnaire | 0.406 | 0.703 | 0.870 | 0.350 | 0.0072 | 7.54× |
| V1 + interview | 0.434 | 0.717 | 0.754 | 0.294 | 0.0076 | 7.14× |
| V2 + inference | 0.434 | 0.717 | 0.826 | 0.313 | 0.0069 | 7.88× |
| **V5 intuitive** | **0.473** | **0.736** | **0.776** | **0.337** | **0.0173** | **3.14×** |
| Human panel | — | — | 0.713 | — | 0.0543 | 1.00× |

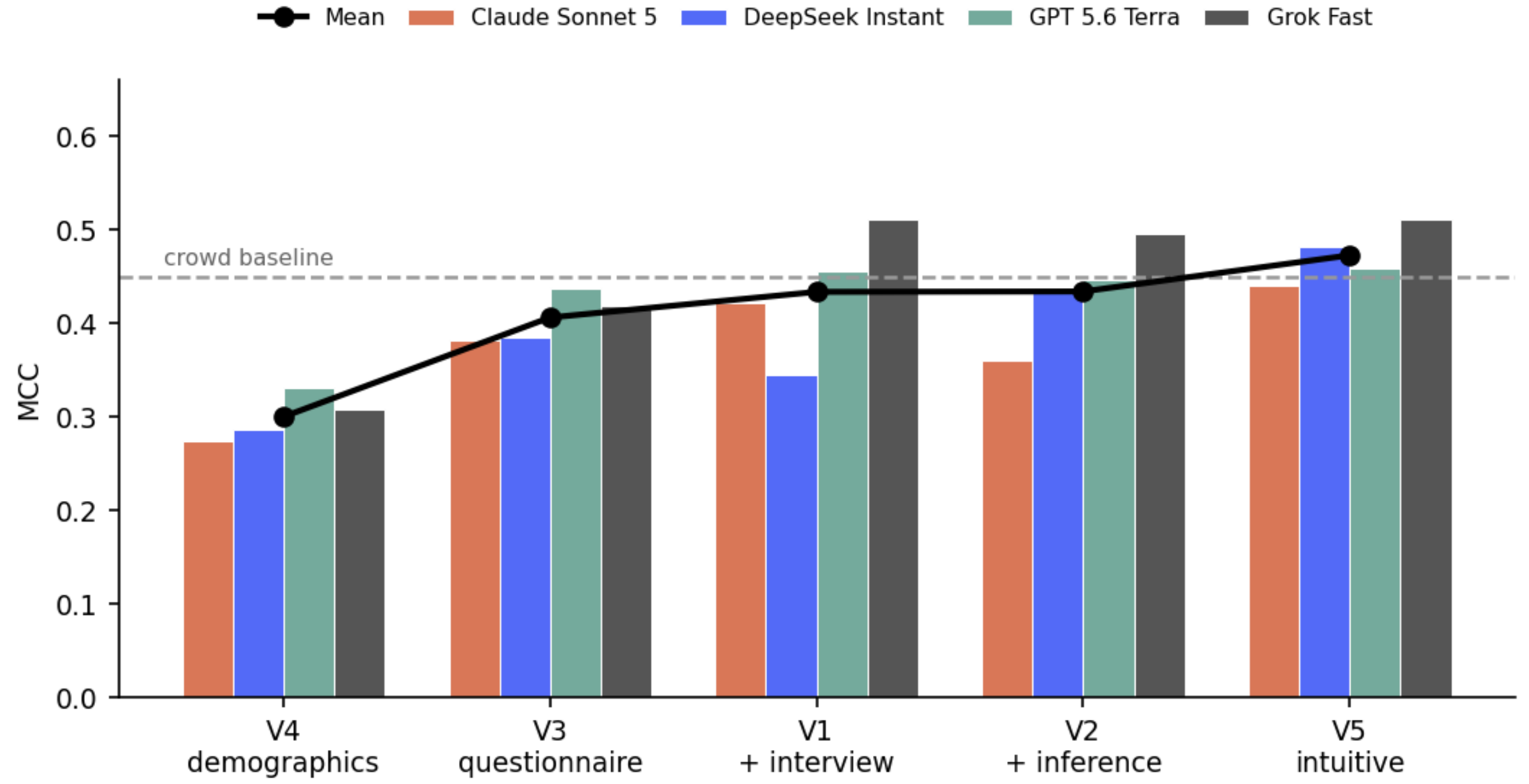

*Figure 2. Fidelity by prompt condition and model, with the crowd baseline shown as a dashed line.*

The step from demographics to the questionnaire is the largest in the ladder and holds for all four models, with 367 item-level cells correct under the questionnaire condition that the demographics condition missed against 252 in the opposite direction ($p < .001$). The comparison supports H2 and speaks directly against the practice of generating synthetic respondents from demographic backstories alone, since knowing a person's age and city buys much less than knowing what that person believes.

Adding the interview transcript and the free-text self-presentation produced a smaller gain of 0.027 in mean fidelity, with 238 cells gained against 208 lost ($p = .170$). The difference reached significance for a single model out of four. The pattern supports H4 in its stated form, which holds that returns to qualitative material are modest once a structured questionnaire is available. Section 3.5 shows that this reading applies to the topics the questionnaire covers and reverses outside them.

Instructing the models to respond immediately rather than analytically produced the highest mean fidelity of any condition and the best result for three models out of four. Item-level comparisons favour the intuitive condition over every alternative, including the richest analytical condition (186 against 144, $p = .024$), the inference condition (143 against 101, $p = .009$), the questionnaire condition (238 against 166, $p < .001$), and the demographics condition (404 against 217, $p < .001$). Treating the model as the unit of analysis leaves the same contrasts short of significance with four models and Wilcoxon p values between .125 and .250, which follows from the sample of models rather than from any weakness in the effect. Both sets of tests appear here because they answer different questions, and the item-level result should be read as the better powered of the two. The evidence supports H7.

The inference instruction failed to improve fidelity. Mean performance under that condition matched the interview condition exactly at 0.434, while profile-consistency rose from 0.754 to 0.826. Telling a model to work out reactions from combinations of characteristics made it follow the profile more closely without bringing it closer to the person, which supports H8 and motivated the intuitive condition that replaced it.

### 3.2 Agents against their own profiles

Profile-consistency was scored for the human panel and for every agent on the same 115 topic cells. Participants matched the pattern implied by their own survey answers in 71.3 per cent of cells under the strict criterion and 79.1 per cent with partial credit. Eleven of the twenty configurations exceeded the human figure, and the three configurations given the questionnaire without narrative material reached 0.904. The finding supports H5.

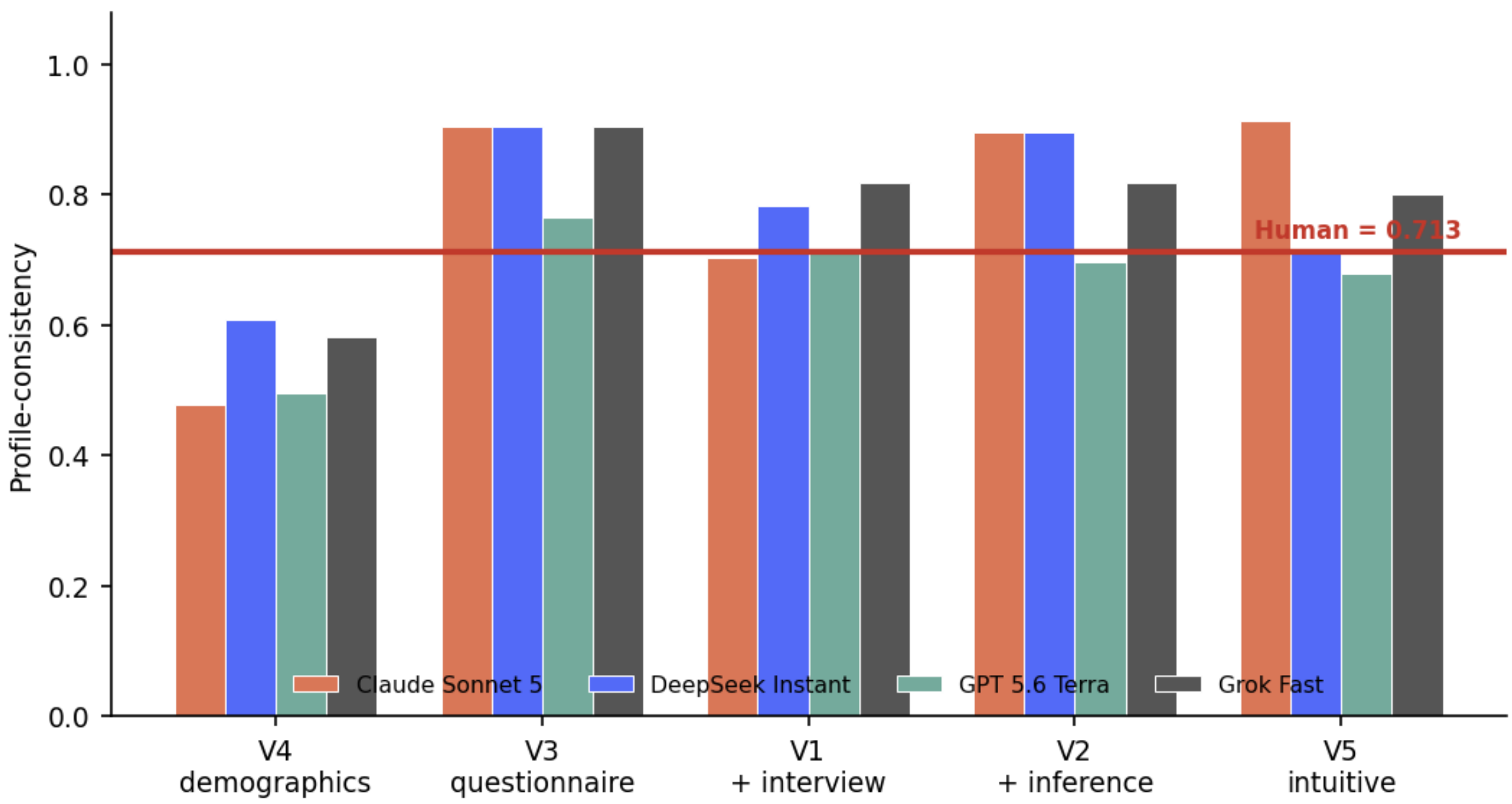

*Figure 3. Profile-consistency by condition and model. The horizontal line marks the value achieved by the human panel.*

The mechanism becomes visible when the conditions are compared. Agents holding a numeric attribute table and little else approach a ceiling near 0.90, reading a trust score from the table and applying it to both posts in a pair. Agents given the interview transcript fall back toward the human level, since narrative material invites judgment about a person rather than retrieval of a value. Human participants sit below every questionnaire-only configuration, which fits a long record of gaps between stated attitudes and observed behaviour and marks perfect consistency as a sign of mechanical operation rather than good simulation.

### 3.3 Individuation

Agents differentiated between participants far less than the participants differed from one another. The human panel showed a between-person standard deviation in like-rate of 0.0543, while the four analytical conditions produced values between 0.0053 and 0.0076, which corresponds to compression by factors of 7.1 to 10.2. Six of the sixteen model-condition cells in those conditions returned a standard deviation of exactly zero, meaning that the model assigned all eight participants an identical overall like-rate.

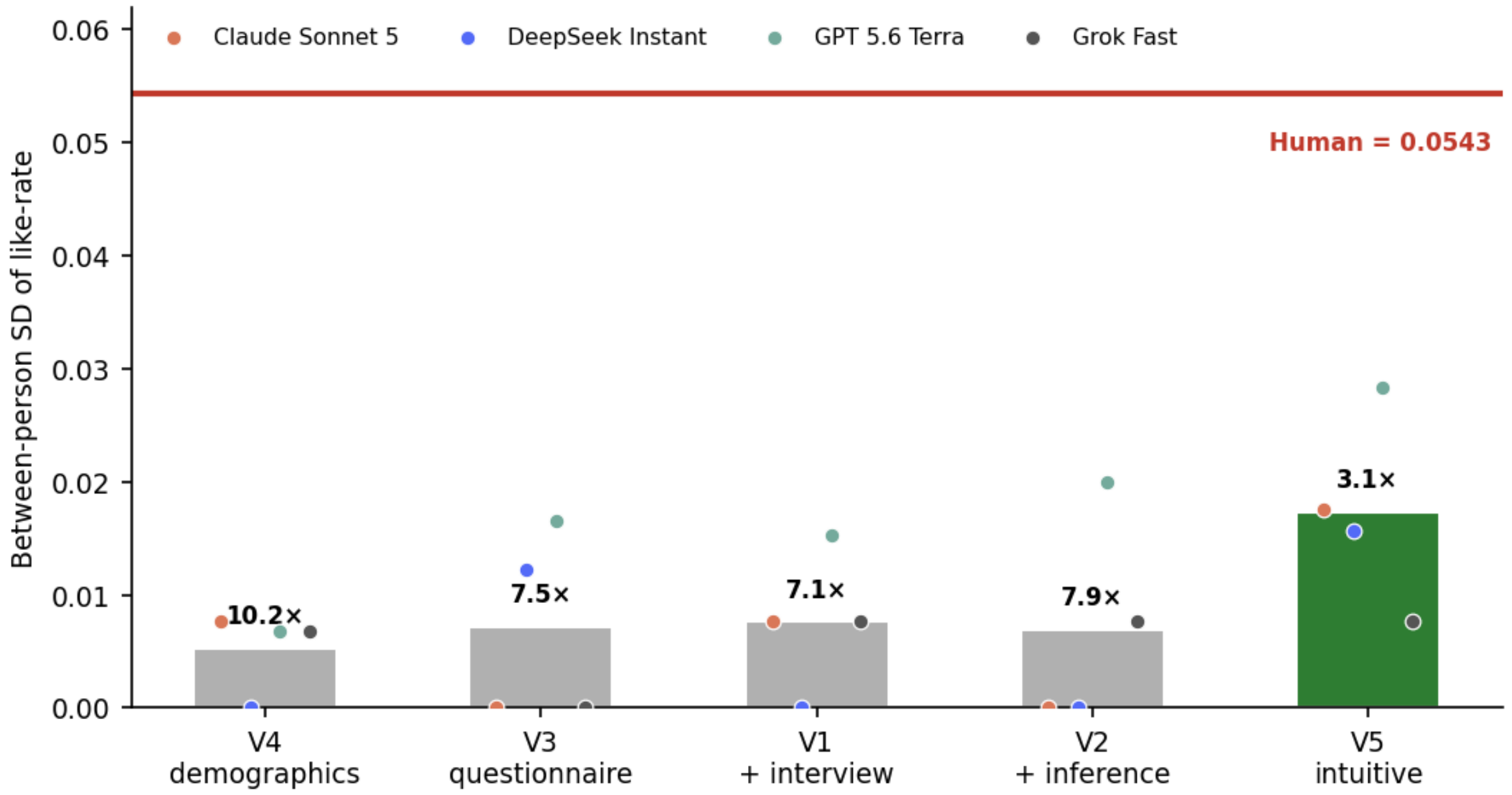

*Figure 4. Between-person standard deviation in predicted like-rate. Bars show the mean across models and points show individual models.*

The intuitive condition changed this picture. Compression fell to 3.1 times the human value, no model returned a standard deviation of zero, and three of the four models improved on their own best result from the analytical conditions while the fourth matched it. The gain in individuation arrived alongside the gain in fidelity, which strengthens the reading of H7 and suggests that deliberation pushes models toward a generic representation of the profile rather than toward the individual it describes.

Estimates of that kind rest on eight numbers, one per participant, so they carry wide uncertainty. Resampling the sixty-eight posts places the difference in between-person standard deviation between the intuitive and interview conditions at 0.004, with a ninety-five per cent interval running from −0.010 to 0.019. All four models moved in the same direction, and three of the four showed a clear gain, so the pattern is consistent across models while the size of the effect stays loosely bounded.

### 3.4 Consistency and fidelity as separate properties

Correlating the two outcomes across all twenty configurations gives a positive association of $r = .490$ ($p = .028$). Restricting the calculation to the sixteen configurations that received attitudinal profile information reverses the sign to $r = -.364$ ($p = .166$). The pattern fits H6 in its non-monotonic form, though the second correlation falls short of significance at this number of configurations.

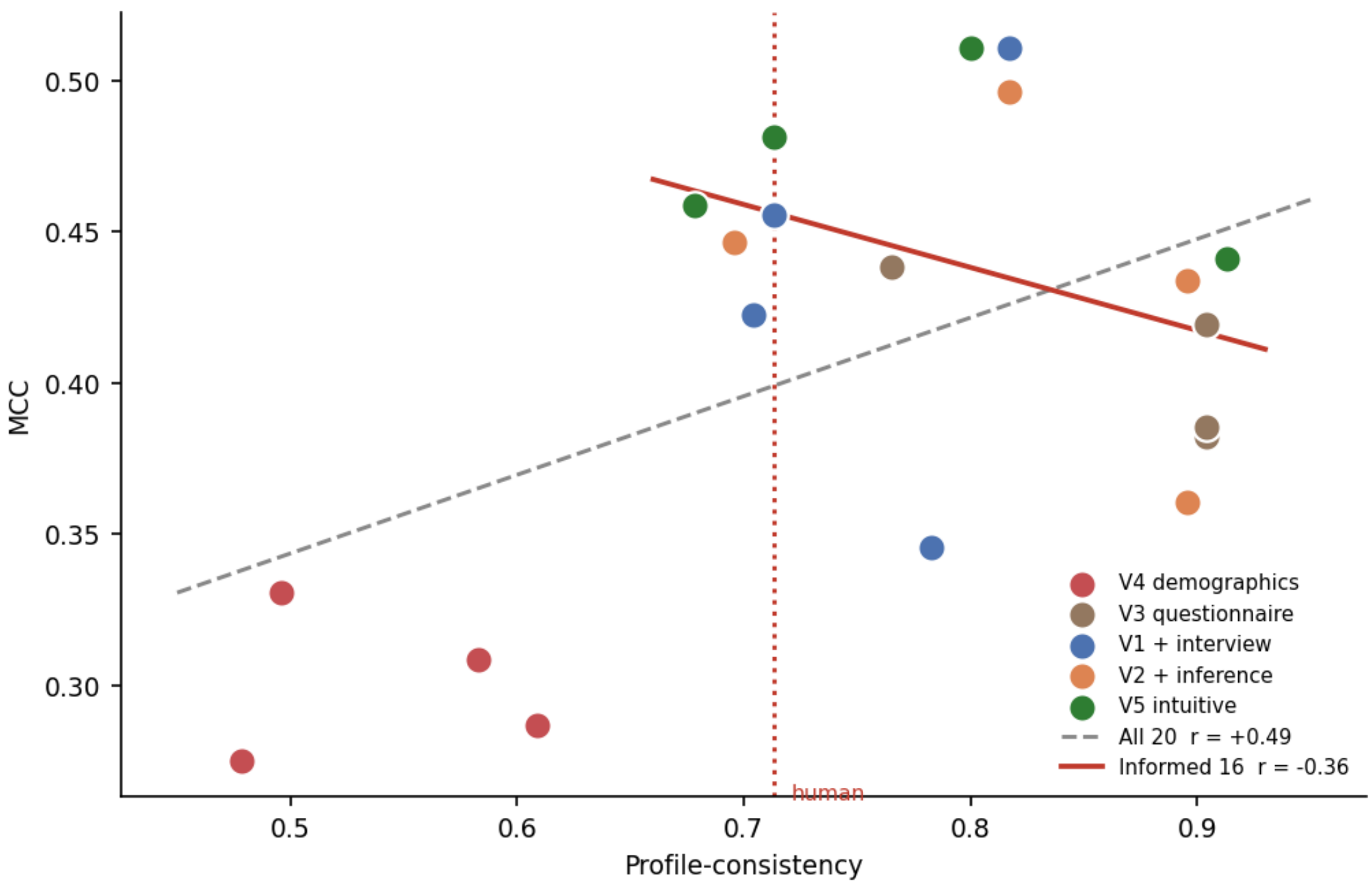

*Figure 5. Profile-consistency plotted against fidelity for all twenty configurations, with separate fitted lines for the full set and for the informed conditions.*

Configurations given demographics alone occupy the lower left corner of Figure 5 and fail on both measures, since an agent without a profile can neither match one nor reproduce the person. The positive association across the full set comes from those four points. Once attitudinal information is present the relationship inverts, with the three most consistent configurations sitting in the middle of the fidelity range while the best performing configurations span consistency values from 0.678 to 0.913. Profile-consistency therefore indicates how mechanically an agent operates rather than how well it represents a person, which recommends its use as a diagnostic alongside fidelity rather than as a target in its own right.

### 3.5 Generalisation to unprofiled topics

Half the stimulus set concerns topics the questionnaire never raised, which provides the test of whether a profiled agent reacts sensibly to unfamiliar content. Fidelity on those posts stayed between 0.324 and 0.399 across conditions, comfortably above the crowd baseline of 0.265 on the same items, with 18 of the 20 configurations exceeding that baseline and the strongest reaching 0.506. The evidence supports H9 in its qualified form. Table 2 reports fidelity for each condition across the post categories.

*Table 2. Fidelity by post category, averaged across the four models.*

| Condition | All posts | Related (34) | Unrelated (34) | News (38) | Entertainment (30) | Positive (31) | Negative (31) |
|---|---|---|---|---|---|---|---|
| V4 demographics | 0.300 | 0.245 | 0.358 | 0.307 | 0.291 | 0.349 | 0.270 |
| V3 questionnaire | 0.406 | 0.489 | 0.324 | 0.468 | 0.329 | 0.473 | 0.366 |
| V1 + interview | 0.434 | 0.495 | 0.374 | 0.477 | 0.380 | 0.491 | 0.402 |
| V2 + inference | 0.434 | 0.513 | 0.357 | 0.488 | 0.368 | 0.470 | 0.417 |

| Condition | All posts | Related (34) | Unrelated (34) | News (38) | Entertainment (30) | Positive (31) | Negative (31) |
|---|---|---|---|---|---|---|---|
| **V5 intuitive** | **0.473** | **0.550** | **0.399** | **0.504** | **0.435** | **0.498** | **0.457** |

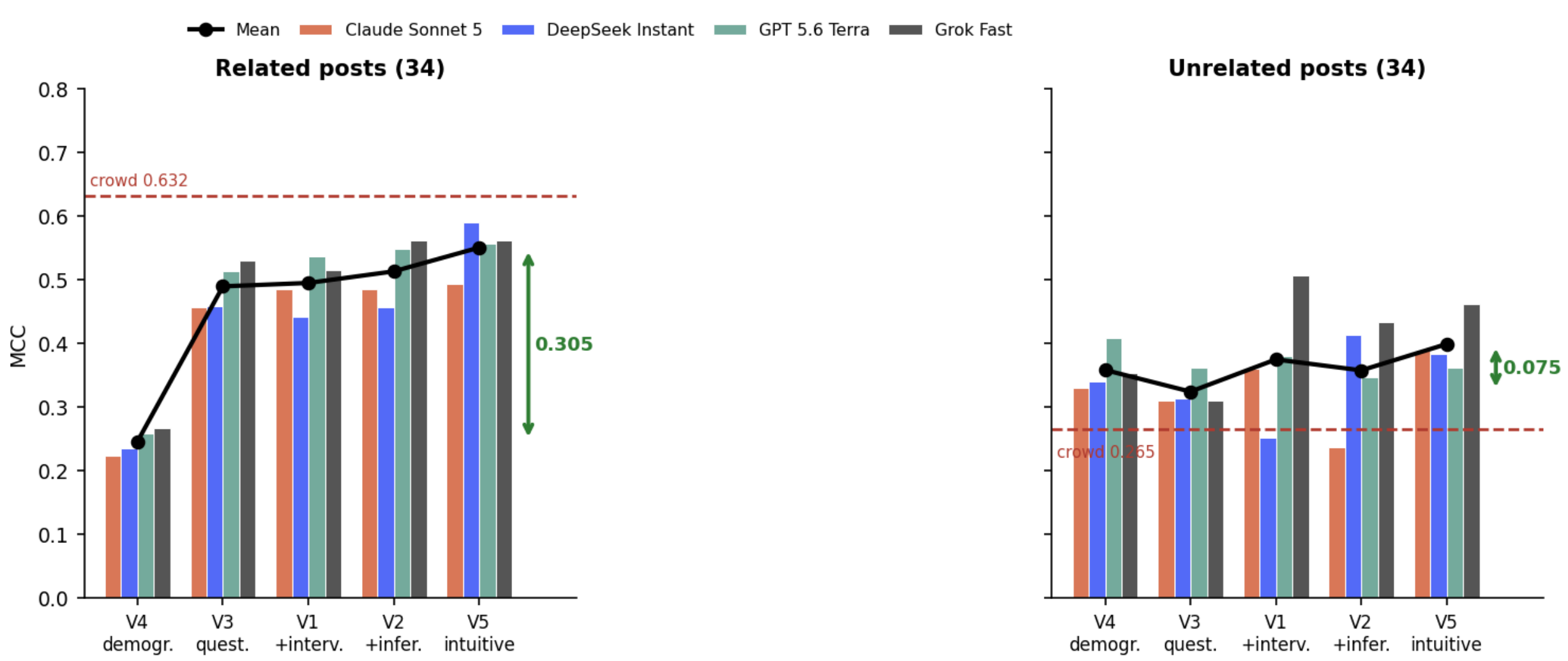

*Figure 6. Fidelity split by whether a post concerns a topic covered by the questionnaire. Green arrows mark the spread between conditions within each panel.*

The spread between conditions collapses once the analysis moves outside the surveyed topics. Fidelity on related posts ranges from 0.245 to 0.550 across the five conditions, a spread of 0.305, while fidelity on unrelated posts ranges from 0.324 to 0.399, a spread of 0.075, with one pairwise contrast reaching significance. A profile therefore raises accuracy most on the topics it covers, while the instruction style still matters outside them, since the intuitive condition beat the questionnaire-only condition on unrelated posts by 161 cells to 120 ($p = .017$).

The apparent advantage of the demographics condition on unrelated posts deserves care, since that condition ranks third of five in absolute terms. Its favourable ratio of unrelated to related performance follows from its collapse on related posts rather than from any strength on unrelated ones. Stereotype matching offers no better explanation, because the demographics condition agreed with the group-majority reaction on unrelated posts in 60.6 per cent of cases, which sits inside the range of 60.9 to 64.3 per cent covered by the other conditions.

Content supplied as narrative rather than as a table proved more useful outside the surveyed topics. The questionnaire-only condition generalised worst of all five at 0.324, falling below the demographics condition. Adding the interview raised transfer to 0.374 and adding intuitive framing raised it to 0.399. The contribution of the interview, worth 0.027 and short of significance within the surveyed topics, rises to 0.050 outside them and separates conditions that transfer from one that does not. The result supports H10 and gives the qualitative material a defensible role that the in-domain comparison alone would have hidden.

The intuitive condition performed best of all five on unprofiled topics, reaching 0.399 against 0.374 for the interview condition and 0.324 for the questionnaire-only condition. Its margin over the crowd baseline on those posts is substantial rather than marginal, since it predicted correctly on 241 cells where the baseline failed against 167 in the opposite direction ($p < .001$). All four models beat that baseline individually on unrelated posts, with scores between 0.361 and 0.462. Individuation followed the same pattern, since compression on unprofiled topics fell to 3.48 times the human level under intuitive framing while the four analytical conditions stayed between 7.1 and 8.0. Agents prompted this way therefore carry individual-level signal into content the profile never described, which is the property a general-purpose simulated user would need. The one contrast that fails to separate is the intuitive condition against demographics alone ($p = .203$), so the questionnaire's marginal value over a demographic sketch stays unproven outside the surveyed topics even though transfer itself is clear.

### 3.6 Comparison against baselines

All baselines were fitted under leave-one-participant-out cross-validation. A random forest trained on the full set of survey variables reached 0.302 and logistic regression reached 0.183, both below every configuration that received attitudinal profile information. The comparison supports H3 and indicates that individual reactions resist recovery from questionnaire variables by conventional supervised learning at this sample size. Table 3 sets the strongest configurations against the full set of baselines.

Table 3. Best-performing configurations against the five baselines.

| Predictor | Accuracy | MCC | Information available |
|---|---|---|---|
| **Grok Fast V1 and V5** | **0.756** | **0.511** | **Full profile** |
| DeepSeek Instant V5 | 0.741 | 0.481 | Full profile, intuitive framing |
| Crowd prior | 0.724 | 0.450 | Reactions of the other seven participants |
| Random forest | 0.651 | 0.302 | All survey variables |
| Logistic regression | 0.592 | 0.183 | All survey variables |
| Majority class | 0.511 | 0.000 | Nothing |
| Participant prior | 0.415 | −0.180 | Own modal reaction |

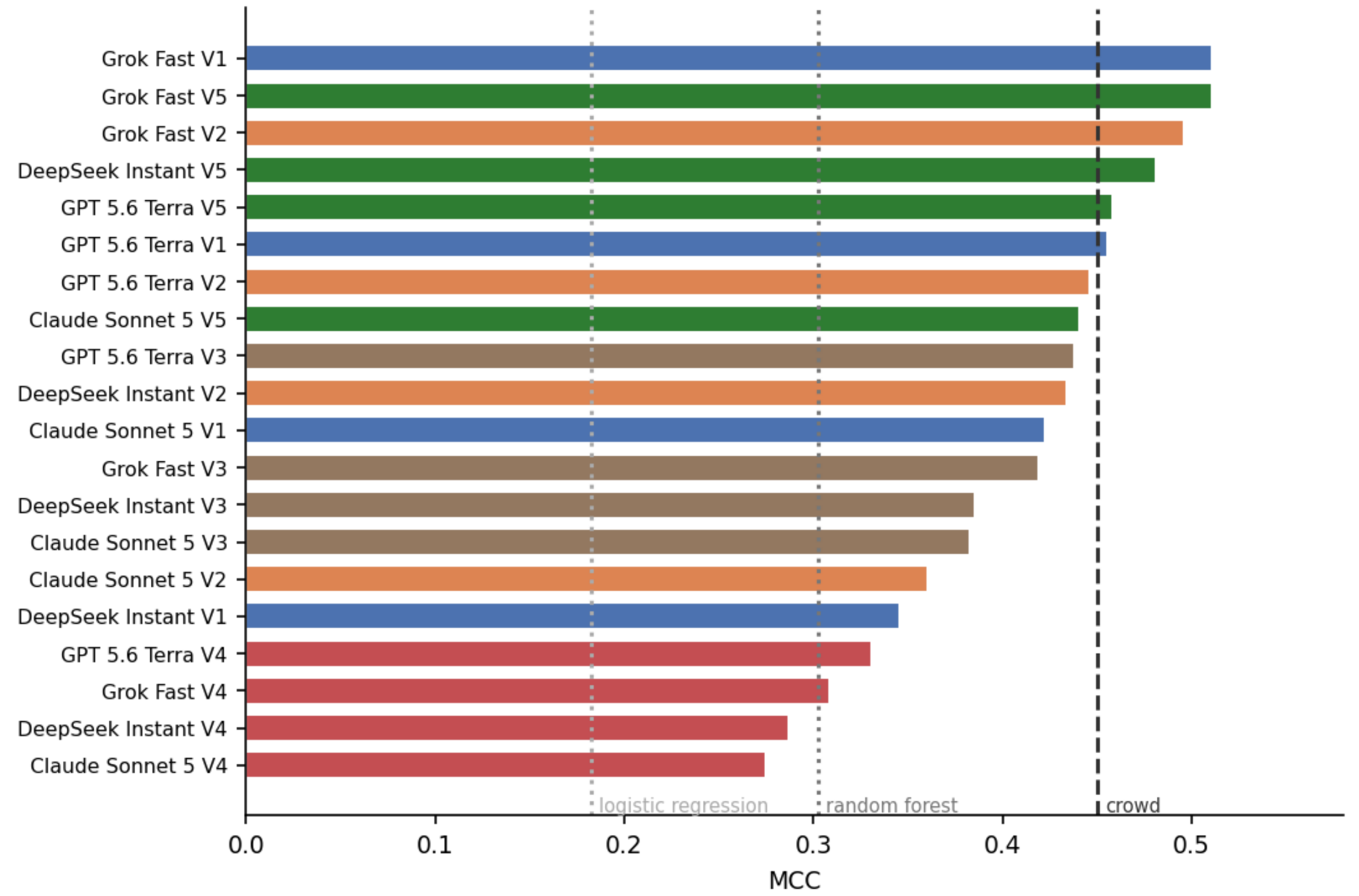

*Figure 7. All twenty configurations ranked by fidelity against the crowd and supervised learning baselines. Colour indicates prompt condition.*

The crowd prior sets the demanding comparison, since predicting each post's modal reaction among the other seven participants reached 0.450 while knowing nothing about the individual. Six configurations beat it outright. Performance of that baseline depends entirely on agreement within the panel, as work on aggregation has long recognised (Surowiecki, 2004). Restricting attention to the 33 contested posts drops the crowd prior to −0.008, which is chance, while every configuration stays positive and the questionnaire and intuitive conditions reach 0.350 and 0.337. Agents therefore carry individual-level signal that aggregation cannot reproduce, and the apparent strength of the crowd prior across the full stimulus set comes from the many posts on which the panel agreed.

Combining the twenty configurations by majority vote gave 0.739 in accuracy, which falls below the best single configuration at 0.756. All twenty configurations were simultaneously wrong on 8.3 per cent of items and simultaneously right on 43.4 per cent. Errors of that kind are shared rather than independent, which limits what ensembling can recover and points to systematic misjudgment of these participants rather than random noise (Kuncheva & Whitaker, 2003).

### 3.7 Statistical considerations

Two units of analysis appear throughout the results. Item-level tests treat each of the 544 agent-post cells as an observation and detect differences between conditions with reasonable power. Model-level tests treat each of the four models as an observation and lack power by construction, since four paired observations cannot reach conventional significance under a Wilcoxon test regardless of effect size. Reporting both keeps the strength of the evidence visible.

The comparisons across conditions form a family of related tests, and the item-level contrasts reported above would survive correction by the Holm procedure (Holm, 1979) at the level of the four ablation contrasts, with the weakest of them close to the threshold. A false discovery rate approach would leave the same conclusions in place (Benjamini & Hochberg, 1995). Effect sizes rather than p values carry most of the argument here, and the differences that matter for interpretation, such as the fall in compression from 7.1 to 3.1, are large by any conventional standard (Cohen, 1988).

Cohen's kappa tracked the Matthews coefficient closely across all configurations, with values between 0.346 and 0.511. The two measures agreed on the ranking of conditions, which is expected given the near-balanced response distribution, though the Matthews coefficient remains preferable as the primary measure because kappa behaves poorly under some marginal distributions (Delgado & Tibau, 2019).

## 4. Conclusion

The study asked how faithfully language model agents reproduce the social media reactions of specific people, and whether the way an agent is asked to use a profile matters as much as what the profile contains. Eight participants were profiled through a questionnaire, a deep interview, and a written self-presentation, then recorded their reactions to sixty-eight posts. Four models predicted those reactions under five prompt conditions, producing twenty configurations and 10,880 predicted reactions. Results are summarised against the ten hypotheses below, followed by what they mean for research practice and where the design falls short.

### 4.1 Answers to the hypotheses

Evidence from the twenty configurations supports nine of the ten hypotheses. Prediction of individual reactions ran above chance throughout, with the Matthews correlation coefficient ranging from 0.275 for Claude Sonnet 5 given demographics alone to 0.511 for Grok Fast under the interview and intuitive conditions, which settles H1. Attitudinal profile content proved far more useful than demographic description, raising mean fidelity from 0.300 to 0.406 with 367 item-level gains against 252 losses ($p < .001$), and H2 holds. Every configuration that received attitudinal information outperformed a random forest at 0.302 and logistic regression at 0.183 under leave-one-participant-out validation, which supports H3 and indicates that individual reactions resist recovery from questionnaire variables by conventional supervised learning at this sample size.

Returns to the qualitative layer were modest within the surveyed topics. Adding the interview transcript and the written self-presentation to the questionnaire gained 0.027 in mean fidelity with 238 gains against 208 losses ($p = .170$), and the difference reached significance for a single model out of four, so H4 holds in the limited form in which it was stated. Human participants matched the pattern implied by their own survey answers in 71.3 per cent of cells, while eleven of the twenty configurations scored higher and the three questionnaire-only configurations reached 0.904, which supports H5 and marks perfect internal coherence as a sign of mechanical operation rather than good simulation.

The relationship between the two outcome measures took the shape predicted by H6. Correlation across all twenty configurations came to $r = .490$ ($p = .028$) and reversed to $r = -.364$ ($p = .166$) once the demographics condition was removed, so profile information is needed for either outcome while greater consistency stops implying greater fidelity once that information is present. Instructions that suppress deliberation produced the highest mean fidelity of any condition at 0.473 and cut compression of individual differences from seven times the human level to three, which supports H7 and gives the study its main constructive result.

Only H8 failed to hold, and its failure was informative. The instruction to infer reactions from combinations of characteristics left mean fidelity unchanged at 0.434 while raising profile-consistency from 0.754 to 0.826, so the models followed the profile more closely without coming closer to the person. That outcome motivated the intuitive condition that replaced it. Transfer to topics the questionnaire never raised stayed between 0.324 and 0.399 against a crowd baseline of 0.265, with 18 of 20 configurations above that baseline while the spread between conditions fell from 0.305 to 0.075, which supports H9 in the qualified form in which it was posed. Content supplied as narrative rather than as a table carried that transfer, since the questionnaire-only condition generalised worst of all five at 0.324 while the interview raised it to 0.374 and intuitive framing to 0.399, and H10 holds. Intuitive framing also gave the best result on unprofiled topics at 0.399, beating the crowd baseline there by 241 cells to 167 ($p < .001$) and the questionnaire-only condition by 161 to 120 ($p = .017$), which supports H11.

Three findings carry the argument. Attitudinal profile content improves individual prediction substantially over demographic backstories, which sets a clear requirement for anyone building synthetic respondents. Profile-consistency and predictive fidelity behave as separate properties once a profile is supplied, so an agent that follows its profile closely can be further from the person than one that departs from it. Instructions that suppress deliberation improve both fidelity and individuation, which shows that the framing of the task is a design variable rather than a detail of implementation.

### 4.2 Implications

The findings speak to the threat raised at the beginning of this paper. Agents built from attitudinal profiles predicted individual reactions at a level well above chance and above what knowledge of the surrounding group could achieve, which means that convincing artificial users of specific kinds are already within reach of anyone with a questionnaire and a public model. Fake accounts of that sort with added memory and specific instruction would gain their power from acting in concert, since a swarm of profiles that each behave like a plausible local citizen can shift the apparent distribution of opinion without any single account looking unusual. The compression reported here works against the deceiver in one respect, because agents that flatten individual differences will produce a crowd that agrees with itself more than a real crowd does, and detection methods might exploit that signature. Prompt design that reduces compression narrows the gap, which places the defensive value of this work alongside its methodological value.

Transfer to unfamiliar content changes what these agents could be used for. An agent that predicts well only on the topics its profile covers is a specialist, useful for studying attitudes that were measured in advance and little else. Results reported here show the intuitive condition holding its advantage on posts about foreign policy, media genres, and sport, none of which the questionnaire raised. A simulated population built this way could be dropped into a feed of arbitrary content and still react in ways that track the individuals it represents. The prospect cuts both ways. Research gains a general-purpose instrument for testing how a population would respond to material that has yet to be written, and anyone building deceptive accounts gains profiles that stay convincing across whatever the news cycle produces. The same property that makes the tool useful makes the threat durable.

Simulated populations offer a way to study questions that cannot be put to real users. Changes to a recommender algorithm cannot be tested on a national electorate without exposing that electorate to the consequences, so a population of agents that reacts to content the way particular people would offers the only practical route to asking how polarisation might develop under a different ranking rule. Work of this kind has already begun, with simulated feeds used to compare alternative news algorithms and to trace opinion dynamics through networks of agents. Results reported here place a condition on that programme, since a simulated public whose members resemble one another will understate disagreement, lose minority positions, and return a comfortable answer about a policy that would divide real people. Compression by a factor of seven would make almost any algorithmic change look safer than it is.

The finding on intuitive prompting raises a question that reaches past this application. Reasoning is treated across most of the field as something to add rather than something to weigh, and models are routinely pushed to deliberate more on the assumption that more thinking yields better answers. Results here suggest a cost that has gone largely unexamined, since the question worth asking is what gets lost when reasoning is switched on, and for which kinds of task. Judgments made quickly and without justification appear to belong to a family where deliberation actively hurts, and the boundaries of that family remain unmapped. The practical value of the finding is more immediate. Agents told to answer at once rather than work through the profile predicted people better than any other condition tested, and they cut the flattening of individual differences from seven times the human level to three. Nothing about the model changed and nothing about the profile changed, because the same participant description produced both results. A few sentences of instruction carried the entire effect, which means the improvement arrives without extra data collection, fine-tuning, or compute. Reactions to social media posts are made in a second or two, and asking a model to reason toward such a judgment appears to move it toward a generic version of the person described rather than toward the person. Matching the manner of the judgment to the manner in which people make it therefore looks as consequential as matching the content of the profile, and the gain from that single choice was larger here than the gain from adding a deep interview to a questionnaire.

Profile-consistency deserves a place in validation practice alongside agreement with human behaviour. The measure separates agents that reason about a person from agents that read values off a table, and the two look identical when only fidelity is reported. Human participants matched their own stated attitudes in 71.3 per cent of cases, which fits a long record showing that people describe their own dispositions imperfectly (Nisbett & Wilson, 1977). An agent scoring far above that level has drifted toward a coherence that human behaviour does not display, so consistency belongs among the diagnostics rather than among the targets.

The value of qualitative profile material depends on where the agent is asked to operate. Interview transcripts and written self-presentations added little within the topics the questionnaire already covered, yet they made the difference between transferring and failing to transfer outside those topics. Research that expects agents to react to novel content should collect narrative material, while research confined to well-specified domains may reasonably rely on structured instruments. The finding gives a concrete answer to a question that qualitative and computational traditions have argued over in general terms (Bail, 2024).

Broader caution follows from the pattern of errors. Combining all twenty configurations by majority vote performed worse than the best single configuration, and every configuration failed simultaneously on 8.3 per cent of items. Shared error of this kind indicates systematic misjudgment of these participants rather than random noise, which limits what more models or larger ensembles can fix. Simulation offers real value for exploring collective patterns while remaining unreliable as a replacement for asking people directly (Crockett & Messeri, 2023; Harding et al., 2024; Lin, 2025), and the risk of mistaking fluent output for understanding is worth keeping in view (Messeri & Crockett, 2024). Careful attention to what these systems represent, and to what they leave out, remains the condition for using them well (Bender et al., 2021).

### 4.3 Limitations and future research

Eight participants support statements about what happens within a rich-profile design. Claims about population parameters would need a different study. Model-level tests across four models cannot reach conventional significance under a Wilcoxon procedure, whatever the size of the effect. Evidence for differences between conditions therefore rests on item-level comparisons that treat the 544 agent-post cells as observations. All participants live in Serbia, while the models tested are known to represent wealthy Western publics more accurately. Performance reported here may understate what the same designs would achieve elsewhere. The consistency measure depends on a rule set linking survey items to expected reactions, and reasonable alternatives exist. Trust scores of four or five were treated as endorsement, and the neutral midpoint was excluded, which removed 21 of 136 topic cells. All thresholds sit in a configuration block so that the sensitivity of the findings to these choices can be checked, which limits the analytic flexibility that undermines replication (Simmons et al., 2011). One post pair concerning home decoration produced human consistency of 0.125. That value looks like a flawed item rather than evidence about the participants. Transfer was tested on a single set of unprofiled topics chosen by the research team, so the generality of that result depends on those topics resembling the content an agent would meet in use. Prompt conditions varied in more than one way at a time. The intuitive instruction combined speed, a ban on justification, and permission to depart from the profile, so which element carries the effect stays open. Each configuration was run once, reactions were reduced to a binary choice, and two models were excluded because their outputs were incomplete.

Replication with a larger panel would settle whether the intuitive advantage holds. A larger panel would also give model-level tests the power they lack here. Running each configuration several times at non-zero temperature would separate the effect of framing from the variability of sampling. Testing more capable models and a wider range of prompt variations would very likely lift performance beyond what is reported here. The four models used were the versions available during data collection, and the five conditions cover a small part of the space of possible instructions. The most valuable target for that search is a prompt formula that reaches high fidelity while keeping consistency close to the human level rather than far above it. Configurations that predicted people best were spread across the consistency range and offered no single recipe. Decomposition of the intuitive instruction would identify which element reduces compression, using separate conditions for speed pressure, for the ban on justification, and for permission to depart from the profile. Comparison against models run with and without explicit reasoning modes would address the same question from the other direction (Binz & Schulz, 2023; Akata et al., 2025). Extension to other countries and languages would show how far the pattern travels. Application of the consistency measure beyond social media reactions would establish whether the ceiling near 0.90 for attribute-only prompting and the human value near 0.71 appear across tasks, or belong to judgments made in a second or two. Prediction of individual behaviour deserves attention as a target in its own right, since psychology has tended to favour explanation over prediction and has paid for that preference in reproducibility (Yarkoni & Westfall, 2017; Open Science Collaboration, 2015). Agents that predict what a particular person will do provide a demanding test of whether a profile captures anything real.

**Ethical Approval**

This study was reviewed and approved by Ethical Committee of the University of Belgrade, Institute for Philosophy and Social Theory on 12 May 2026. All procedures performed in this study involving human participants were in accordance with the ethical standards of the institutional research committee and with the 1964 Helsinki declaration and its later amendments or comparable ethical standards.

**Informed Consent:**

Informed consent was obtained from all individual participants included in the study. Prior to participation, participants were provided with detailed information about the study's purpose, procedures, potential risks, and benefits. They were assured of the confidentiality and anonymity of their responses and that their participation was voluntary. Participants indicated their consent by clicking a consent button.

**Competing Interests:**

The authors declare that they have no competing interests.

**Author Contributions:**

LB: Conceptualization, Methodology, Formal analysis, Writing - Original Draft, Visualization.
TS: Investigation, Resources, Data Curation, Writing - Original Draft.
JM: Conceptualization, Methodology, Formal analysis, Writing - Original Draft, Visualization.
ADS:: Conceptualization, Writing - Original Draft.
BD: Conceptualization, Writing - Original Draft.
JW: Conceptualization, Writing - Original Draft.

**Data Availability**

The data that support the findings of this study are available from the Open Science Framework (OSF). The dataset is openly accessible at
https://osf.io/gdq9j/overview?view_only=4a9096e0201a487c8c9231d148293dda

**Acknowledgements**

This research was supported by the Joint Excellence in Science and Humanities (JESH) programme awarded by the Austrian Academy of Sciences (OeAW).

This paper was realised with the support of the Ministry of Science, Technological Development and Innovation of the Republic of Serbia, according to the Agreement on the realisation and financing of scientific research.